\documentclass[journal]{IEEEtran}

\usepackage{amsmath,amssymb,amsfonts}
\usepackage{algorithmic}
\usepackage{graphicx}
\usepackage{textcomp}
\usepackage{xcolor}
\usepackage[ruled,vlined]{algorithm2e}
\usepackage{hyperref}
\usepackage{optidef}
\usepackage{multirow}
\usepackage{hyperref}
\usepackage{xcolor}
\usepackage{soul}
\usepackage{subfig}
\usepackage{physics}
\usepackage{comment}
\def\BibTeX{{\rm B\kern-.05em{\sc i\kern-.025em b}\kern-.08em
    T\kern-.1667em\lower.7ex\hbox{E}\kern-.125emX}}
\newcommand{\wprime}{\textbf{w'}}

\newcommand{\rpm}{\raisebox{.2ex}{$\scriptstyle\pm$}}

\makeatletter
\def\footnoterule{\kern-3\p@
  \hrule \@width 2in \kern 2.6\p@} 
\makeatother

\hypersetup{
    colorlinks=true,
    citecolor=black,
    linkcolor=black,
    filecolor=magenta,      
    urlcolor=blue,
}

\begin{document}

\title{AdaSwarm: Augmenting Gradient-Based Optimizers in Deep learning with Swarm Intelligence}

\author{Rohan~Mohapatra,
        Snehanshu~Saha,~\IEEEmembership{Senior Member,~IEEE,}
        ~Carlos~A.~Coello~Coello,~\IEEEmembership{Fellow,~IEEE,}
        ~Anwesh~Bhattacharya,
        ~Soma~S.~Dhavala,
        ~and~Sriparna~Saha, ~\IEEEmembership{Senior Member,~IEEE}
\thanks{Manuscript received September 27, 2020; revised February 6, 2021; accepted April 26, 2021. Date of current version April XX, 2021.}

\thanks{R. Mohapatra is with the Center for AstroInformatics (Astrirg), Bengaluru, India, e-mail: \texttt{rohannmohapatra@gmail.com}. S. Saha is with the Dept. of Computer Science and Information Systems and APPCAIR, BITS Pilani, K K Birla Goa, Goa Campus, India, e-mail: \texttt{snehanshu.saha@ieee.org}.} 

\thanks{C. A. C. Coello is with the Department of Computer Science, CINVESTAV-IPN, 	Mexico City, Mexico, Basque Center for Applied Mathematics (BCAM) \& Ikerbasque, Spain, e-mail: \texttt{ccoello@cs.cinvestav.mx}. A. Bhattacharya is with the Department of Physics and CSIS, BITS Pilani, Pilani, India, e-mail: \texttt{f2016590@pilani.bits-pilani.ac.in}.} 

\thanks{S. S. Dhavala is with ML Square, Bengaluru, India, e-mail: \texttt{soma@mlsquare.org}. S. Saha is with Department of Computer Science and Engineering, Indian Institute of Technology Patna, India, e-mail: \texttt{sriparna@iitp.ac.in}}

\thanks{Digital Object Identifier }

\thanks{© 20XX IEEE. Personal use of this material is permitted. Permission from IEEE must be obtained for all other uses, in any current or future media, including reprinting/republishing this material for advertising or promotional purposes, creating new collective works, for resale or redistribution to servers or lists, or reuse of any copyrighted component of this work in other works.}

}

\markboth{IEEE Transactions on Emerging Topics in Computational Intelligence,~Vol.~X, No.~Y, Month~2020}%
{Mohapatra \MakeLowercase{\textit{et al.}}: AdaSwarm: Augmenting gradient-based optimizers in Deep learning with Swarm Intelligence}

\maketitle

\begin{abstract}
    This paper introduces AdaSwarm,  a novel gradient-free optimizer which has similar or even better performance than the Adam optimizer adopted in neural networks. In order to support our proposed AdaSwarm, a novel Exponentially weighted Momentum Particle Swarm Optimizer (EMPSO), is proposed. The ability of AdaSwarm to tackle optimization problems is attributed to its capability to perform good gradient approximations. We show that, the gradient of any function, differentiable or not, can be approximated by using the parameters of EMPSO. This is a novel technique to simulate GD which lies at the boundary between numerical methods and swarm intelligence. Mathematical proofs of the gradient approximation produced are also provided. AdaSwarm competes closely with several state-of-the-art (SOTA) optimizers.  
    We also show that AdaSwarm is able to handle a variety of loss functions during backpropagation, including the maximum absolute error (MAE).
\end{abstract}


\IEEEpeerreviewmaketitle

\section{Introduction}
Gradient Descent (GD) \cite{Rumelhart1986} is the most popularly used update rule, utilized heavily to update the weights associated with different layers when errors are back-propagated in shallow and deep neural networks (DNN). It is a numerical approximation method embraced in training neural networks (NN). GD is a dependable and potent method, primarily based on derivative-based optimization. The dependence on derivatives implies it is applicable only on functions differentiable everywhere in their domain. Computation of derivatives for functions endowed with complex behavior, often encountered in real-world scenarios, can be arduous. To overcome the drawbacks of GD, metaheuristic optimization based algorithms have been developed. Particle swarm optimization (PSO) \cite{realPSOPaper}, originally proposed in 1995, is a prominent member in the family of such metaheuristic algorithms. In an attempt to model the behaviour of bird flocks, PSO was first proposed by Kennedy and
Eberhart. It is a gradient free optimizer in comparison to other optimization techniques such as GD. 

PSO has been used in a wide variety of real-world applications, including: engineering optimization problems \cite{gargGaPso,springerBook}, search for habitable exoplanets 
\cite{theo2018,BORA2016129,BASAK2020100335}, power system operation
 \cite{khesti2018}, and generation of stable structures of carbon clusters \cite{fchem}, among many others. PSO has also been adopted in several NN applications \cite{Garro2015, jha2009,rauf2018, 8063385}, in which it has been used to optimize and generate weights with high accuracy and low loss. 
For example, Grimaldi et al. \cite{Grimaldi} introduced techniques to train a NN with PSO. They conceptualized the particles comprising the weights of the NN and devised a way of reducing the loss in training by an equivalence conjecture between the fitness function and the loss function. While this approach is reasonable to apply in NN with limited number of weights, the particle dimension is linear in the number of layers of the network. If we consider a real-world application of a Convolutional Neural Network (CNN) \cite{Fukushima1980, recentadvance}, the number of weights would be in millions. In such large-scale problems, PSO is not expected to be competitive with respect to classical optimizers. However, instead of particles being weights, a theoretical correspondence between PSO and the error gradient, supported by empirical validation, will enable us to mitigate the curse of dimensionality. This is equivalent to establishing facts mathematically that the gradient of a function can be approximated from the knowledge of the values of the hyperparameters of PSO. This will allow us to part away with actual derivative computation, especially for functions with jagged landscapes. Using the approximate gradients, we can  then emulate gradient optimization techniques and apply in different types of NNs.
In this regard, AdaSwarm is proposed to show that a ``gradient''-free optimizer\footnote{Gradient-free in this context means using the approximate gradient values instead of the analytically derived values} can provide results that are competitive with respect to the best optimizers currently available.

\subsection{Research Background}
\label{section:psoVariants}

\subsubsection{PSO with Inertial Weight}

The primary objective of PSO was to solve optimization problems with several objectives (\emph{devoid of exact, analytical optima}). PSO navigates the search space by embracing a {\em swarm} which is nothing but a population of particles. The swarm, guided by characteristic equations, attempt to converge to an optima \cite{realPSOPaper}. The movement of the particles in the search space to discover the optimal solution is governed by the velocity and the position update equations. These are represented in the following manner:  
\begin{align}
    v_i^{(t+1)} &= \phi v_i^{(t)} + c_1r_1(p_i^{best} - x_i^{(t)}) + c_2r_2(g^{best} - x_i^{(t)})\label{eq:weightedPSO} \\
    x_i^{(t+1)} &= x_i^{t} + v_i^{(t+1)}\label{eq:2}
\end{align}
where \begin{math}\phi, c_1, c_2\geq0\end{math}. Here, \begin{math}x_i^{t}\end{math} suggests the position of particle $i$ at time $t$, \begin{math}v_i^{t}\end{math} represents the velocity of particle $i$ at time $t$, \begin{math}p_i^{best}\end{math} is the best position particle $i$ has attained, and \begin{math}g_i^{best}\end{math} is the best position that the swarm has ever attained. Additionally, parameters related to cognitive learning and social learning need to be defined, which regulate the position and velocity of the particles and they are $c_1$, $c_2$ respectively; $r_1$, $r_2$ are random values sampled from $U(0,1)$ which contributes to the stochastic nature of the search process.

\subsubsection{PSO with Momentum}
Backpropagation algorithms often use a momentum term to steer away from stagnating local minima. This is typically done by introducing a momentum term. Inspired by this, the velocity update equation of PSO can accommodate momentum to handle the stagnation problem. The standard PSO equations are embellished with the momentum term, and the resulting algorithm is called MPSO\cite{momentumPSO}. 
\begin{align}
    v_i^{(t+1)} &= \left[v_i^{(t)} + c_1r_1(p_i^{best} - x_i^{(t)}) + c_2r_2(g^{best} - x_i^{(t)})\right]\times \nonumber \\
    & (1-\lambda) + \lambda v_i^{t-1}
\end{align}
Here, $\lambda$ denotes the momentum factor.


\subsection{Motivation}

The sources of inspiration to understand and improve optimization techniques in general, and PSO in particular, come from several places. Said et al. \cite{said} postulated that swarms behave similarly to classical and quantum particles. In fact, their analogy is so striking that one may 
think that the social and individual intelligence components in PSO are, after all, nice useful metaphors, and that there is a neat underlying dynamical system at play. 
This dynamical system perspective was indeed useful in unifying two almost parallel streams, namely, optimization and Markov Chain Monte Carlo sampling  \cite{teh,jordan}. 
In a seminal paper, Wellington and Teh \cite{teh}, showed that a stochastic gradient descent (SGD) optimization technique can be turned into a sampling technique by just adding noise, governed by Langevin dynamics. Recently, Soma and Sato \cite{ito} provided further insights into this connection based on an underlying dynamical system governed by stochastic differential equations (SDE). While these results are new, the connections between derivative-free optimization techniques based on Stochastic Approximation and Finite Differences are well documented \cite{spall}. Such strong connections between these seemingly different subfields of optimization and sampling made us wonder: Is there a larger, more general template of which the aforementioned approaches are special cases?
SGD (SGD) \cite{robbins1951} was proposed to tackle the slow speed of convergence of the GD method and also to improve the minimization of the underlying optimization problem. However, this approach is not good enough in many cases. 
What are we missing and what is the role of metaheuristics in this framework?
This particular question gives rise to two issues: can we use ideas developed for improving SGD (SGD) and apply them to PSO? Also, can we use the particles' history and their location awareness to offer derivative-free approximations to the gradients? In this paper, we answer these questions in the positive. 

\subsection{Summary of Contributions}
\textbf{Section~\ref{section:empso}} proposes EMPSO, a PSO variant aided with exponentially averaged momentum,  contributing to reduced number of iterations required to minimize the error and reach the optima for a variety of benchmark (\emph{differentiable and non-differentiable}) test functions (\textbf{Section ~\ref{section:equivalence}}, Tables ~\ref{table:gdVsEmulatedGD}, ~\ref{table:sgdVsEmulatedSGDEMPSO} and ~\ref{table:sgdVsEmulatedSGDEMPSONonDiff}) and for several standard single-objective test functions used for global optimization (see the supp file\footnote{supp file = Supplementary Material}, Section 8). EMPSO is markedly different from the existing M-PSO approach (\textbf{Section~\ref{section:empso}} \emph{where the exploration and the exploitation terms are used in an equal manner}). Section~\ref{section:equivalence} presents the mathematical formulation of EMPSO as well as a proof of the theorems establishing mathematical equivalence between Vanilla PSO \& GD (\textbf{Section ~\ref{section:equivalenceGDPSO}}) ---, and between EMPSO \& SGD with Momentum (\textbf{Section~\ref{section:equivalenceSGDEMPSO}}).
Such a theoretical equivalence complements our empirical results on the application of EMPSO to both differentiable and non-differentiable test functions (\textbf{Section ~\ref{section:equivalence}, Tables ~\ref{table:gdVsEmulatedGD}, ~\ref{table:sgdVsEmulatedSGDEMPSO} and ~\ref{table:sgdVsEmulatedSGDEMPSONonDiff})} as well as to standard single-objective test functions (\emph{see the supp file, Section 8}). The Theorems establish a direct gradient approximation method with precise alternatives to gradient computation by exploiting the hyperparameters of the Vanilla and EMPSO.
Since our proposed approach is an approximation method, we discuss its order of accuracy in the same section briefly and present a proof for it (\emph{see Section 1.1 in the supp file}). Next, we interpret and extend the gradient approximation approach in \textbf{Section ~\ref{section:interpretGradientsInNN}} to emulate backpropagation in Vanilla feed-forward NN and CNN via EMPSO, tested on a simulated data set\footnote{Link for simulated data set: \url{https://gist.github.com/rohanmohapatra/4e7bce4f0d95746d8b993437c257e99b}} (\emph{see} \hyperref[section:interpretGradientsInNN]{\textbf{Section IV} \emph{and} Fig.~\ref{fig:trueVsApproxGradient}}). 
Results in \textbf{Sections ~\ref{section:equivalence} and ~\ref{section:interpretGradientsInNN}} pave the foundation for a novel adaptive gradient-free  optimizer, which we call AdaSwarm. This approach is intended for optimizing NNs (\textbf{Section~\ref{adaswarm}}) and we show that it produces promising results for multiple classification data sets (\textbf{Tables ~\ref{table:adaswarmClassificationMSELoss} and ~\ref{table:adaswarmClassificationMAELoss} in the paper and Table 5} in the supp file). \textbf{Section ~\ref{section:experiments}} reviews differentiable and non-differentiable loss functions adopted in NNs. In order to facilitate training using high dimensional classification data, we present a rotational variant of EMPSO, called Rotation Accelerated EMPSO (REMPSO)  (\emph{see Section 5 of the supp file}) which is intended for computer vision data sets (\emph{see Section 3 of the supp file}). \textbf{Section ~\ref{section:experiments}} contains the subtle technical aspects of AdaSwarm and reports the results of our experiments on several test functions commonly used in  optimization \cite{Jamil_2013} as well as on data sets used in complex classification tasks in deep learning. We conclude in \textbf{Section ~\ref{section:discussions}} by highlighting possible extensions to AdaSwarm.


\section{Our Contribution: Exponentially weighted Momentum PSO (EMPSO)}
\label{section:empso}

We begin by proposing a novel variant, Exponentially weighted Momentum PSO \cite{rmohapatraempso}, to momentum PSO (MPSO) by assigning greater emphasis to the exploration phase of PSO. The computed weighted average in Momentum particle swarm optimization (M-PSO) algorithm \cite{momentumPSO} contributes to exploration and exploitation simultaneously. Since finding the optima in the search space via PSO is tied to exploration mostly, the exploration part of the PSO equation would benefit if greater weights are assigned to it. 

We formulate a strategy to mitigate the issues MPSO faces by presenting a variant of Particle Swarm Optimizer with momentum. The mathematical representation of EMPSO is  as follows:
\begin{equation}
\label{eq:proposedMPSO}
    v_i^{(t+1)} = M_i^{(t+1)} + c_1r_1(p_i^{best} - x_i^{(t)}) + c_2r_2(g^{best} - x_i^{(t)})
\end{equation}
where,
\begin{equation}
\label{eq:proposedMomentum}
    M_i^{(t+1)} = \beta M_i^{t} + (1 - \beta)v_i^{(t)}
\end{equation}
We use $\beta$ to indicate the momentum factor, and $M_i^{t}$ is the momentum of a particle. The combination of \eqref{eq:proposedMPSO} and \eqref{eq:proposedMomentum} yields a new representation of the mathematical formulation of EMPSO.
\begin{equation}
     v_i^{(t+1)} = \beta M_i^{t} + (1 - \beta)v_i^{(t)} + c_1r_1(p_i^{best} - x_i^{(t)}) + c_2r_2(g^{best} - x_i^{(t)})
\end{equation}

Exploration and exploitation are two important phases for any optimization algorithm. Exploration is done by the algorithm to search for other possible solutions apart from the current solution in all of the search space of the problem (\emph{i.e., finding diverse solutions is preferred at early stages of the search}). On the other hand, exploitation is to search for similar and better solutions around the current solution for higher precision (\emph{i.e., preferable when the search is close to finish}). 
PSO, as any other metaheuristic, operates with these two phases
\cite{PSOThreshold}: 
\begin{equation*}
     \underbrace{v_i^{(t)}}_\text{Exploration}  + \underbrace{c_1 r_1 (p_i^{best} - x_i^{(t)}) + c_2 r_2 (g^{best} - x_i^{(t)})}_\text{Exploitation}
\end{equation*}
The exploration phase, due to the new approach, is now determined by the \textbf{exponential weighted average of the historical velocities} only. The negligible weights in MPSO do not contribute to the required acceleration \cite{momentumPSO}. The momentum in EMPSO is the exponential collection of velocities that the particles have seen so far, over time. We accumulate the velocities by multiplying by an exponential factor $\beta$, as we move ahead in time. Thus, $\beta$ factor assigns more weight to the recent velocity than to older velocities.

\subsection{Intuition behind Exponentially Weighted Average}

Let us recursively expand \eqref{eq:proposedMomentum} for further intuition on exponentially averaged momentum.  Let's begin with $ M_i^{(t)} = \beta M_i^{(t-1)} + (1 - \beta)v_i^{(t-1)}$ which is expanded easily to $M_i^{(t)} = \beta^2 M_i^{t-2} + \beta(1 - \beta)v_i^{t-2} + (1 - \beta)v_i^{t-1}$. Generalizing --- 
\begin{multline}
    M_i^{(t)} = \beta^n M_i^{(t-n)} + \beta^{(n-1)}(1 - \beta)v_i^{(t-n)} + \\
    ... +\beta(1 - \beta)v_i^{(t-2)} + (1 - \beta)v_i^{(t-1)}
\end{multline}
From the above equation, we can see that the value of the $t^{th}$ term of the momentum is dependent on all the previous values of the velocities. The $(t - i)^{th}$ time step is assigned the weight $\beta^i (1- \beta)$. Since $\beta<1$, we have $\beta^{i+1}<\beta^i$and the associated weights decrease rapidly. Thus, older velocities get a much smaller weight and, therefore, contribute less to the overall value of the Momentum. 
Since the velocities are a cumulative sum, storing the velocity history of a particle is not required.

\subsection{EMPSO Algorithm}
\begin{algorithm}[h]
\SetAlgoLined

\SetKwFor{For}{for}{do}{endfor}
 \For{each particle i in swarm S}{
    initialize particle with feasible random position\;
    evaluate the fitness $F_i$ of particle\;
 }
 \While{termination condition is not fulfilled}{
    \For{each particle i in swarm S}{
       $ v_i = \beta M_i + (1 - \beta)v_i + c_1r_1(p_i^{best} - x_i) + c_2r_2(g^{best} - x_i)$\;
       $ x_i = x_i + v_i$\;
       $ M_i = \beta M_i + (1 - \beta)v_i$\;
    }
    update $p_i^{best}$, $g^{best}$ by finding the best fitness\;
    choose best solution\;
 }
 \caption{EMPSO}
\end{algorithm}

The parameters adopted are the following:
\begin{itemize}
    \item $S$: Swarm containing a certain number of particles
    \item $x_i$: Position of Particle \textit{i}, $\beta$: Momentum factor
    \item $v_i, M_i$: Velocity/Momentum of Particle \textit{i}
    \item $c_1, c_2$: Cognitive/Social Learning Parameter
    \item $r_1, r_2$: Random values between [0,1]
    \item $p_i^{best}$: Personal best velocity of Particle \textit{i}
    \item $g^{best}$: Global best velocity observed in Swarm $S$
\end{itemize}
\textbf{Complexity Analysis of EMPSO:}
The EMPSO algorithm has a complexity of $\mathcal{O}(t \times n \times log(n))$. Considering the worst case of finding the maximum, EMPSO at the worst case has a computational complexity $\mathcal{O}(t\times n^2)$. This computational cost is relatively low because it is linear in $t$. The number of particles, $n$ is normally a relatively small number in PSO (\emph{See Section 7 of the supp file for detailed calculations}).



\section{Approximation of Gradients: Differentiable/Non-Differentiable Functions}
\label{section:equivalence}
We present three theorems establishing approximate gradients for any arbitrary (\emph{loss}) function, dependent on the PSO/EMPSO parameters. The theorems thus could be applied to either differentiable or non-differentiable loss functions in training DNNs.

\subsection{Proof of Equivalence between the GD and a Vanilla PSO  to approximate gradients for Differentiable functions}
\label{section:equivalenceGDPSO}
\noindent \textbf{Theorem} - Under reasonable assumptions, for an arbitrary objective function \begin{math}f(w)\end{math} (\emph{differentiable or not}), if PSO converges at \begin{math}w=g^{best}\end{math}, the gradient approximation theorem states that \begin{math}\pdv{f}{w}=\frac{-(c_1r_1 + c_2r_2)}{\eta}(w-g^{best})\end{math}

\noindent \textbf{Proof} - First consider a single-dimensional objective $f(x)$. One can expand it as a Taylor-series about $x=a$ as 
\begin{multline}
    f(x) = f(a) + f'(a)(x-a) + \frac{f''(a)}{2}(x-a)^2 +\\ \ldots + E_n(x;a) \label{eq:obj.taylor}
\end{multline}
where $|E_n(x;a)|=\frac{k(x-a)^{n+1}}{(n+1)!}$ is the error of the approximation. We can compute the first derivative as ---
\begin{multline}
    f'(x)=f'(a)+f''(a)(x-a)+\ldots+E_{n-1}(x;a) \label{eq:objder.taylor}
\end{multline}
Analogously, for a multidimensional objective $f(w)$, centering its Taylor-expansion about $w=\textbf{\wprime}$, one can express ---
\begin{multline}
     \pdv{f}{w}=f'(\wprime) + f''(\wprime)(w-\wprime)\\ + \ldots + E_{n-1}(w;\wprime) \label{eq:objparder.taylor}
\end{multline}
Strictly speaking, $f'=\nabla_wf$ and $f'' = \pdv{f}{w_i}{w_j}$ with $w_i$ ranging over the dimensions of the weights. The slight abuse of notation lends to better understanding of the theorem. The weight update rule of GD (\emph{used in backpropagation to train NNs}) is\footnote{See section (\ref{sec:backprop}) for a derivation} ---
\begin{align}
w^{(t+1)} = w^{(t)} + \eta\eval{\pdv{f}{w}}_{w^{(t)}} \label{eq:grad.desc}
\end{align}
At the timestep with weight $w^{(t)}$, the Taylor-expansion of the derivative (eq \ref{eq:objparder.taylor}) [$w^{(t)}$ is the weight matrix of the NN, $\eta$ is the learning rate and $f$ is the loss function for the given network] could be substituted in the GD update and written as ---
\begin{multline}
     w^{(t+1)} - w^{(t)} = \eta[f'(\wprime) + f''(\wprime)(w^{(t)}-\wprime) \\+ \ldots  + E_{n-1}(w^{(t)};\wprime)] \label{eq:objparder.grad.desc}
\end{multline}
The purpose of GD is to find the minimum of an objective. However, the same minima could be found by PSO if we let $x\equiv w$ i.e., the particle dimensions are that of the weights of the NN. The PSO position update equation could be put into the form ---
\begin{align}
    & x^{(t+1)} - x^{(t)} \nonumber \\
    &= v^{(t+1)} \nonumber \\
    &= \omega v^{(t)} + c_1r_1(p^{best} - x^{(t)}) + c_2r_2(g^{best} - x^{(t)}) \label{eq:obj.pso}
\end{align}
For the corresponding PSO search process with $x^{(t)}$, the particle trajectory, prior to convergence, takes vanishingly small steps and $v^{(t)}\approx 0$. The swarm satisfies $p^{best}\approx g^{best}$ near convergence \cite{realPSOPaper}. The point $\wprime$, about which the Taylor expansion is done, is constructed such that $g^{best}=\wprime$ after convergence. The error term $E_{n-1}(w;\wprime)$ could be ignored for practical purposes. Moreover, if $\wprime$ is a minima $\implies$ $f'(\wprime)=0$. We make the following assumptions before asserting the equivalence ---
\begin{enumerate}
    \item $w^{(t)}$ is inside a small, $\delta-$neighborhood centred about $\wprime$
    \item $\eta = \omega$ for mathematical convenience (\emph{not required for the minima condition, near or far from the minima in a NN setting, $f'(\wprime)=0$ irrespective of this setting})
    \item $x\equiv w$ i.e., $x, w$ are used interchangeably; $x$ in PSO is used to update particle position while $w$ in a neural-network setting, is used to update weights.
\end{enumerate}
Under the above assumptions and settings, it is possible to equate the RHSs of eq (\ref{eq:objparder.grad.desc}) and (\ref{eq:obj.pso}) to obtain :
\begin{align*}
    \eta f''(\wprime)(w^{(t)}-\wprime) &= c_1r_1(p^{best} - x^{(t)}) +  c_2r_2(g^{best} - x^{(t)}) \\
    \implies f''(\wprime)(w^{(t)}-\wprime) &= -\frac{(c_1r_1 + c_2r_2)(w^{(t)} - g^{best})}{\eta}
\end{align*}
since $p^{best} \approx g^{best}$, $x\equiv w$ and $g^{best}\approx \wprime$. Thus, we obtain
\begin{equation}
    f''(\wprime) = -\frac{(c_1r_1 + c_2r_2)}{\eta};  
    \label{eq:grad.equiv.dder}
\end{equation}     
Upon substituting eq (\ref{eq:grad.equiv.dder}) back into eq (\ref{eq:objder.taylor}) along with $f'(\wprime)=0$ and $g^{best}\approx \wprime$, we finally arrive at the gradient equivalence theorem ---
\begin{align}
    \pdv{f}{w} = \frac{-(c_1r_1+c_2r_2)}{\eta}(w-g^{best}) \label{eq:grad.equiv.der}
\end{align}
Instead of the true gradient, it is now possible to use the approximated gradient evaluated at $w^{(t)}$ for running GD. This is called gradient-descent with PSO approximated gradient. 

It is possible that assumption 1 above does not hold. In such cases where we are searching the search space and the points are farther away from the \textbf{optima}, the error will be larger. The $p^{best}$ will factor in for points away from the minimum and hence the equivalence can also be written as ---
\begin{itemize}
    \item $ v^{(t)}_i = f'(\textbf{w}') + E_{n-1}(w)$
    \item $f''(\wprime) = -\frac{\left[c_1r_1(w^{(t)}-p^{best}) + c_2r_2(w^{(t)}-g^{best})\right]}{\eta(w^{(t)}-\wprime)}$
    \item $\frac{\partial f}{ \partial w} = v^{(t)}_i - \frac{c_1 r_1(w^{(t)} - p^{best}_i) +c_2 r_2(w^{(t)} - g^{best})}{\eta} $
\end{itemize}
Next, we extend the equivalence to a more advanced version of the GD that provides a faster convergence and avoids stagnation at local minima.


\subsection{Proof of Equivalence between the SGD with Momentum and EMPSO to approximate gradients for Differentiable (loss) functions}
\label{section:equivalenceSGDEMPSO}
\textbf{Theorem:} Under reasonable assumptions in the neighborhood of the minima, the following equivalence holds:
$\eta = (1- \beta)$,  $f''(\textbf{\wprime}) =  \frac{-(c_1 r_1 + c_2 r_2)}{\eta}$ and $\alpha = \beta $\\
\\
\textbf{Proof:} Please see Section 1 in the supp file for details.

\subsection{A Subtlety Regarding Non-Differentiable Loss Functions}
\label{section:equivalenceSGDEMPSONonDiff}

Let's assume a continuous, bounded function $f (x)$ that is not differentiable. It can be expressed as $f (x + \alpha)$, where  $\alpha$ is a shift parameter with an initial value equals to zero. Let us also define $z \equiv x +  \alpha$, so that $f(z) \equiv f (x +  \alpha)$. To define the Taylor series for such a function, $f$ must be  differentiable with respect to  $\alpha$, and consequently $f$ must be differentiable with respect to $z$. When a function is shifted by a parameter $\alpha$, we can see that the function becomes differentiable in that interval. Alghalith \cite{taylor_non_diff} introduced differentiability with respect to a shift parameter rather than the variable because a small change in the shift parameter is a constant. Since $f$ is differentiable with respect to $z$, the Taylor series expansion around $a$ can we written as: 
\begin{equation*}
    f(z) = f(a) + f'(a)(z-a) + \frac{f''(a)}{2!} (z-a)^2 + .... + E_n (z, a)
\end{equation*}

Let's consider a non-differentiable function, $f(x)$. If it is expressed by using a shift-parameter $\alpha$, then replacing $x+\alpha$ by $z$, we can make the function differentiable under $z$ (\emph{as every function, differentiable or not can be shifted}). For details, please refer to Moawia Algalith's paper \cite{taylor_non_diff}. The taylor series expansion, in $z$, remains the same as before. Then, the gradient of this non-differentiable function can be expressed with PSO parameters in the same way as for a differentiable function.
\begin{equation*}
\label{}
    f'(z) = 0 + f'(a) + f''(a)(z-a) + \frac{f'''(a)}{2!} (z-a)^2 + .... + E_{n-1} (z, a)
\end{equation*}
The GD Weight Update with momentum is given by:
\begin{equation*}
    w^{(t+1)} = w^{(t)} + \eta V_{dw}^{t+1}
\end{equation*}
where,
\begin{equation}
\label{eq:GradientMomentumNonDiff}
    V_{dw}^{t+1} = \beta V_{dw}^{t} + (1-\beta) \frac{\partial f}{\partial w}
\end{equation}
Combining the equations and dividing equation~\eqref{eq:GradientMomentumNonDiff} by $(1-\beta)$ we obtain: $w^{(t)} = w^{(t-1)} + \alpha V_{dw}^{t-1} + \eta \frac{\partial f}{\partial w}$

Here, $\eta$ is the learning rate and $V_{dw}^{t-1}$ is the momentum applied to the weight update, where, $ \alpha = \frac{\eta \beta}{(1-\beta)}$

We apply the Taylor series expansion for non-differentiable functions to the gradient and then formulate the GD with momentum as follows:
\begin{multline}
     w^{(t)} = w^{(t-1)} + \eta f'(\textbf{w}') + \eta f''(\textbf{w}')(z-\textbf{w}') \\
     +  E_{n-1} (z) + \alpha V_{dw}^{t-1}
\end{multline}
Under the same assumptions stated in the previous proofs, we define the equivalence as follows:
    $\eta = (1- \beta); \alpha = \frac{\eta \beta}{(1-\beta)}$    
\begin{equation}
\label{eq:fdoubledashderviative}
   f''(\textbf{w}') = -\frac{(c_1 r_1 + c_2 r_2)}{\eta}
\end{equation}
Finally, the gradient for non-difererentiable functions can be expressed as follows:

\begin{equation*}
    \frac{\partial f}{ \partial w} = \frac{-(c_1 r_1 + c_2 r_2)}{\eta}(z - g^{best})
\end{equation*}
Section 1.2 in the supp file for a more complete treatment of this case.

\begin{table*}[htbp]
\begin{center}
\caption{\small{\textbf{Differentiable Functions} : GD \emph{vs.} PSO Emulated GD using $\eta=0.1, c_1=0.8, c_2=0.9$}}
\scalebox{0.8}{\begin{tabular}{|c|c|c|c|c|c|}
\hline
\textbf{Function} & \textbf{Global Minimum} & \textbf{GD Optimum} & \textbf{Iterations} & \textbf{Emulated GD} & \textbf{Iterations} \\ \hline
$-\left(3x^{5\ }-\ x^{10}\right)$ & -2.25 & 2.6 & 31 & -2.249 & 9 \\ \hline
$x^{3}-3(x^{2})+7$ & 3 & 3.00 & 17 & 3.00 & 11 \\ \hline
$-\exp\left(\cos\left(x^{2}\right)\right)+x^{2}$ & -2.718 & -2.718 & 349 & -2.718 & 9 \\ \hline
$x^{15}-\sin\left(x\right)+\exp\left(x^{6}\right)$ & 0.4747 & 2.6 & 548 & 0.4747 & 11 \\ \hline
\end{tabular}}
\label{table:gdVsEmulatedGD}

\caption{\small{\textbf{Differentiable Functions}: SGD with Momentum \emph{vs.} EMPSO Emulated GD $\eta=0.1, c_1=0.8, c_2=0.9, \beta=0.9$}}
\scalebox{0.8}{\begin{tabular}{|c|c|c|c|c|c|}
\hline
\textbf{Function} & \textbf{Global Minimum} & \textbf{SGD Optimum} & \textbf{Iterations} & \textbf{Emulated GD} & \textbf{Iterations} \\ \hline
$-\left(3x^{5\ }-\ x^{10}\right)$ & -2.25 & -1.45 & 667 & -2.249 & 13 \\ \hline
$x^{3}-3(x^{2})+7$ & 3 & 3.0 & 71 & 3.00 & 16 \\ \hline
$-\exp\left(\cos\left(x^{2}\right)\right)+x^{2}$ & -2.718 & -2.718 & 49 & -2.706 &  8\\ \hline
$x^{15}-\sin\left(x\right)+\exp\left(x^{6}\right)$ & 0.4747 & 0.679 & 66 & 0.4747 & 6 \\ \hline
\end{tabular}}
\label{table:sgdVsEmulatedSGDEMPSO}

\caption{\small{\textbf{Non-differentiable Functions} : SGD with Momentum \emph{vs.} EMPSO Emulated GD using $\eta=0.1, c_1=0.8, c_2=0.9, \beta=0.9$}}
\scalebox{0.8}{
\begin{tabular}{|c|c|c|c|c|c|}
\hline
\textbf{Function} & \textbf{Global Minimum} & \textbf{SGD Optimum} & \textbf{Iterations} & \textbf{Emulated GD} & \textbf{Iterations} \\ \hline
$\sqrt x$ & 0 & NA & NA & 0.01 & 22 \\ \hline
$|x|$ & 0 & 0.1 & 1001 & 0.03 & 21 \\ \hline
$\begin{cases} 
    x^{\frac{1}{3}} + sin(x) & x>0 \\
    0 & x\leq 0
\end{cases}
$ & 0 & 0.14 & 13 & 0.004 &  21\\ \hline
$\begin{cases} 
    x & x>0 \\
    x^2 & x<0\\
    0 & x= 0
\end{cases}
$ & 0 & 0.001 & 50 & 0.04 & 20 \\ \hline
\end{tabular}
}
\label{table:sgdVsEmulatedSGDEMPSONonDiff}
\end{center}
\end{table*}


\begin{figure}%
    \centering
    \subfloat[\centering True Gradient \emph{vs.} Approximate Gradient calculated using PSO Parameters: The approximation accomplished is evidence of the efficacy of EMPSO and of the order of accuracy proof presented in Section ~\ref{section:orderOfAccuracy}]
    {{\includegraphics[keepaspectratio=true,scale=0.04]{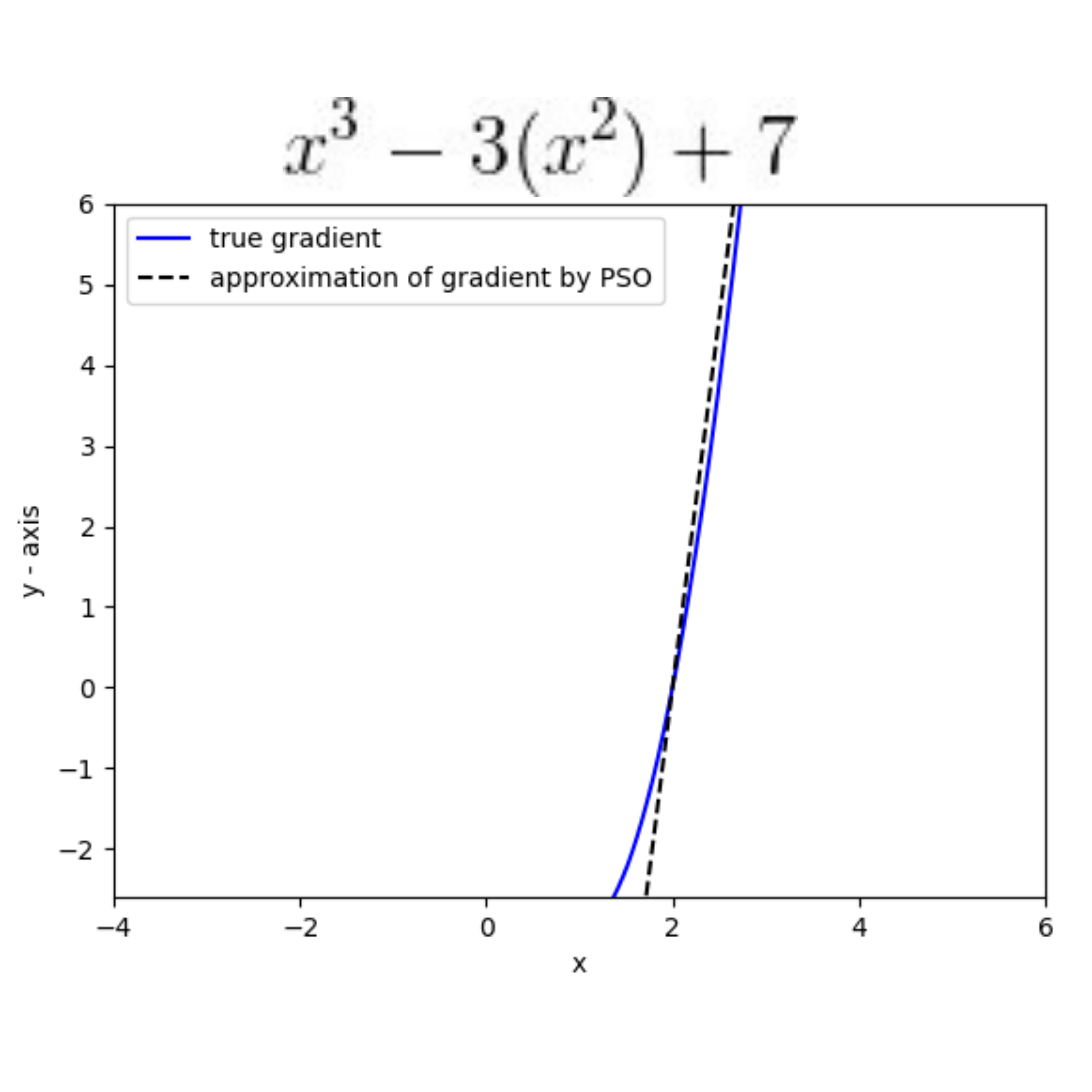} \label{fig:trueVsApproxGradient} }}%
    \qquad
    \subfloat[\centering True Gradient \emph{vs.} Approximate Gradient calculated using PSO Parameters for non-differentiable functions]
    {{\includegraphics[keepaspectratio=true,scale=0.04]{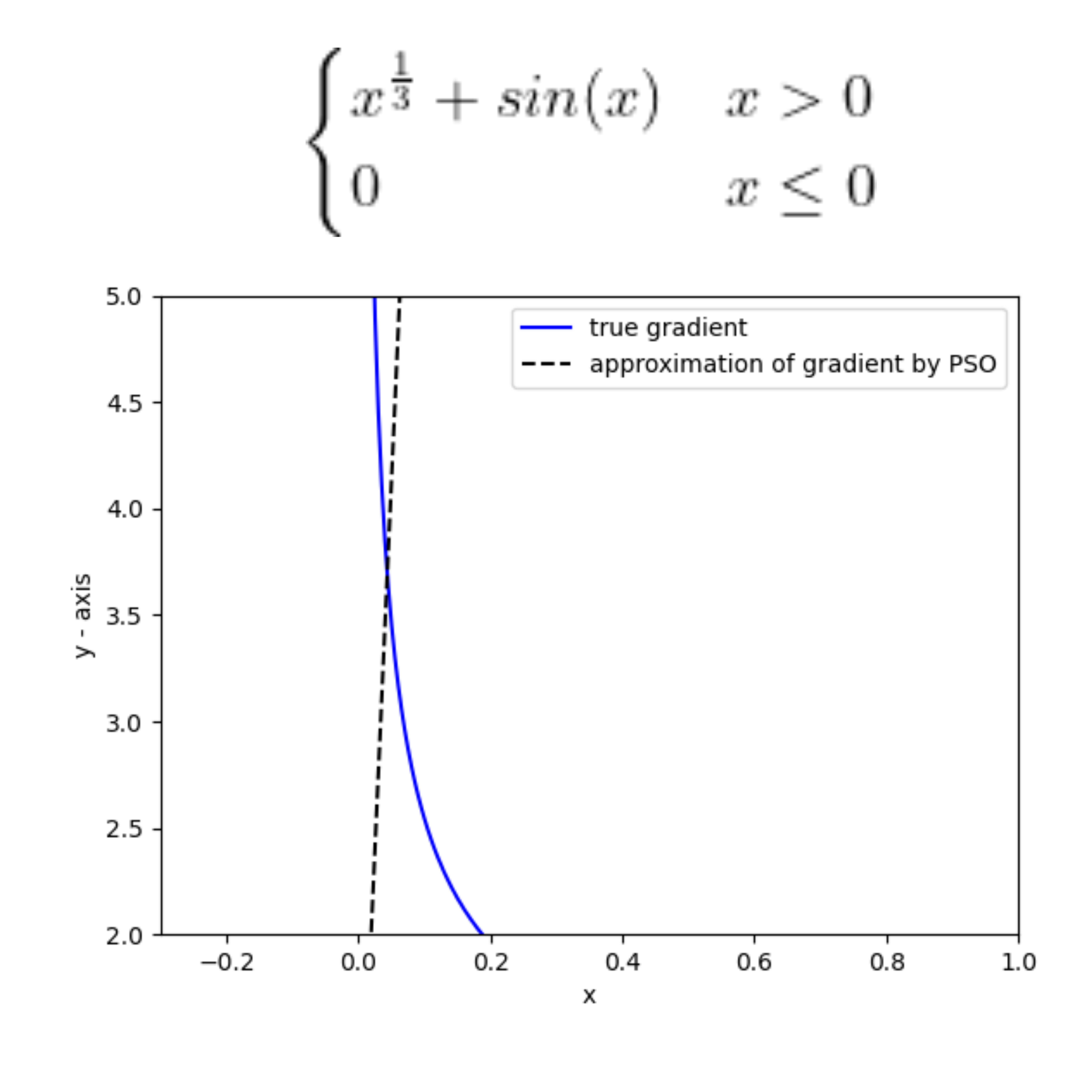}\label{fig:trueVsApproxGradientNonDiff} }}%
    
    \caption{True Gradient \emph{vs.} Approximate Gradient}%
    \label{fig:trueVsApproxGradientFigs}%
\end{figure}

\subsection{Order of Accuracy: EMPSO's gradient approximation}
\label{section:orderOfAccuracy}
The approximation of the exact gradient value depends on a small parameter $h$ (\emph{say the grid size or time step}). 
The order of accuracy (See the proof in Section 1.1 of the supp file) was found to be higher near the optimum. 
We consider two types of functions: smooth functions (\emph{differentiable everywhere}) and non-differentiable functions (\emph{finite number of singularities}). We compute approximate gradients and evaluate approximate optima of these functions using EMPSO. The promising results from this approximation, as theoretically proved, are experimentally validated. We observed a reasonable degree of accuracy between the optima and the approximations produced by EMPSO. The approximation holds for points away from the optima (\textbf{See Table 4 in supp file}). The graphs (see Figure~\ref{fig:trueVsApproxGradientFigs}) support our assertion. The numerical results, tabulated in Tables ~\ref{table:gdVsEmulatedGD}, \ref{table:sgdVsEmulatedSGDEMPSO}, \ref{table:sgdVsEmulatedSGDEMPSONonDiff} 
reflect this observation. 
These experiments suggest the need for investigating the possibility of exploiting mathematical equivalence of gradients in the backpropagation algorithm in the context of DNNs (Sections ~\ref{section:interpretGradientsInNN}-~\ref{section:discussions}).


\section{Interpretation of Gradients and Emulating Back propagation in NNs}
\label{section:interpretGradientsInNN}


Let us revisit the GD rule proposed in the backpropagation algorithm for vanilla feed forward networks:
\begin{equation*}
    w^{(t+1)} = w^{(t)} + \eta \frac{\partial f}{ \partial w}
\end{equation*}

The equivalence between GD (GD) and Particle Swarm Optimization (PSO) has been proved in the previous section. We can now substitute the gradient with PSO parameters and approximate the gradient values. Computing the gradient for some functions might be a very difficult task and some loss functions are non-differentiable. If we can get a good approximation, then we can easily emulate the GD.

Let's consider at any iteration, that the $c_1r_1$ and $c_2r_2$ values are computed by taking the average of $n$ particles, and let's consider the velocity $v^{(t)}_i$  of the particle that influences the $g^{best}$ whose position is $x^{(t)}$  

Using the PSO parameters, we get the following equation:
\begin{multline}
    w^{(t+1)} = w^{(t)} + \eta[v^{(t)}_i - 
    \\ \frac{c_1r_1(x^{(t)}-p^{best}) + c_2r_2(x^{(t)}-g^{best})}{\eta}]
\end{multline}

Tables ~\ref{table:gdVsEmulatedGD}, \ref{table:sgdVsEmulatedSGDEMPSO}, \ref{table:sgdVsEmulatedSGDEMPSONonDiff}  and Fig. ~\ref{fig:trueVsApproxGradientFigs} show the emulation using approximate gradients. For training a NN, PSO has been extensively used \cite{jha2009,rauf2018}, since the GD has a higher chance of getting stuck in local minima. 
PSO's particles encode the weights of the NN and the corresponding fitness function is the loss function that has to be reduced. 
The caveat in this approach is that as the number of weights increase, and so will the number of dimensions of a single particle. Thus, PSO will eventually become a victim of the ``curse of  dimensionality``, since it will fail to converge when dealing with a very large number of weight updates. 

We propose a novel idea to train the NN. We use the concept of \textbf{batch} training and the dimensionality of the particles vector is now just \textbf{ batch\_size $\times$ no\_of\_classes}, thus significantly reducing the amount of computations and increasing the training speed with respect to previous approaches. In a NN, we essentially have a loss function that has to be minimized in order to obtain the set of weights required to classify a given instance. Traditionally, the backpropagation algorithm would do an update as follows: for different loss functions,
the Error Gradient would look like (where $y$ and $\hat{y}$ are the predicted label and the true label, respectively): 
\begin{itemize}
    \item Mean Square Error (MSE) Loss: $\frac{1}{2}(y - \hat{y})^2$
    \begin{equation}
        \frac{\partial E}{\partial w} = (y - \hat{y}) \times D(activation) \times x
    \end{equation}
    \item Binary Cross Entropy (BCE) Loss: $-ylog(\hat{y}) - (1-y)log(1-\hat{y})$
    \begin{equation}
        \frac{\partial E}{\partial w} = \left(\frac{1-\hat{y}}{1-y} - \frac{\hat{y}}{y}\right) \times D(activation) \times x
    \end{equation}
    \item Mean Absolute Error (MAE) Loss: $|y - \hat{y}|$
    \begin{equation}
        \frac{\partial E}{\partial w} = \begin{cases} 
                                           +1 \times D(activation) \times x & y > \hat{y} \\
                                            -1 \times D(activation) \times x & y < \hat{y}
                                       \end{cases}
    \end{equation}
\end{itemize}
 The common equation in all of these losses, is that the neural update rule is given by: 
 \begin{equation}
     \frac{\partial E}{\partial w} = \frac{\partial E}{ \partial y} \times D(activation) \times x
 \end{equation}
 
\par The equivalence between the error gradient and the EMPSO approximation is  natural for any loss function but it's a non-trivial consequence of the insights gained by the theorems proved in 
Section ~\ref{section:equivalence}. If a derivative can be expressed mathematically using PSO and the proposed EMPSO parameters, it is expected that the error gradient whose minimum is critical to compute in the backpropagation algorithm, can be approximated using PSO's parameters. This is a clear departure from existing heuristic-based approaches. 

\begin{figure}[!ht]
\begin{center}
    \includegraphics[scale=0.287]{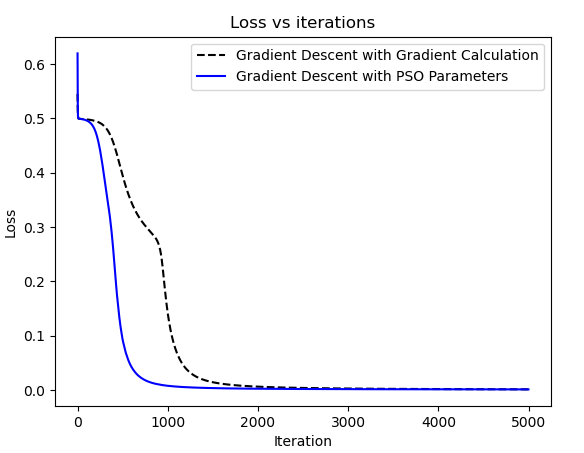}\\
    \caption{Error Gradient approximation in backpropagation by EMPSO on simulated data, as proposed in Eq.~\eqref{eqn:nnequivalence}: 
    }
    \label{fig:errorGradient}
    \end{center}
    \end{figure}

If we can replace $\frac{\partial E}{ \partial y}$ by an approximate value of the gradient, then we can use this gradient for any loss function (\emph{differentiable/non-differentiable}).  Using the theorems, we can then adopt the following weight update rule in NNs:
 \begin{equation}
 \label{eqn:nnequivalence}
    \frac{\partial E}{\partial w} = \frac{-(c_1r_1 + c_2r_2)}{\eta} (g^{best} - y) \times D(activation) \times x
\end{equation}

where $D(activation)$ is the derivative of the activation function adopted. Using this approximation, we are able to descend to the minimum at a faster rate. 
From Fig.~\ref{fig:errorGradient} it can be clearly seen that the descent in the slope 
is steeper than when using the traditional gradient approach.

For MAE, using this loss function we can clearly see that when $y = \hat{y}$, then the MAE derivative is undefined, but when we replace the gradient with the approximate gradients, we do not have to worry about $y = \hat{y}$. In this case, the approximate gradient would be zero and will not update the weight, which is expected. Therefore, we can say that the gradient can be approximated for any differentiable/non-differentiable loss function. This allows the use of this technique in each and every case.

\subsection{Backpropagation and Error Gradient Equivalence in Vanilla Feed-Forward NNs}
\label{sec:backprop}

Consider a node in the final layer of a NN. It has inputs $x_i$ which are the activations of the nodes of the previous layer. The term $net_j$ is a linear sum $net_j=\sum_i w_{ji}x_i + b_j$, and $y_j$ is the result of a univariate activation function $\sigma$, acting on $net_j$, leading to $y_j = \sigma(net_j)$. E is a multivariate loss function with $y_j$ as one of its inputs. Applying the chain rule of derivatives, we get ---
\begin{align*}
    \pdv{E}{w_{ji}} &= \pdv{E}{y_j}\pdv{y_j}{w_{ji}} \\
    &= \pdv{E}{y_j}\pdv{y_j}{net_j}\pdv{net_j}{w_{ji}} \\
    &= \pdv{E}{y_j}\pdv{y_j}{net_j}x_i
\end{align*}
where $y_j$ is the output activation. By approximating $\pdv{E}{y_j}$ with EMPSO, where $E$ could be a non-differentiable loss function, we have the following equivalence (\emph{the detailed proof can be found in Section 2 the supp file}):
\begin{equation}
    \label{backPropTuningRule}
    \frac{\partial E}{\partial y_j } = \frac{-(c_1 r_1 + c_2 r_2)}{\eta}
\end{equation}
This relation is loss function-agnostic. For background on the backpropagation equations and the notation used, please refer \cite{swarthmore}.

\begin{table*}[htbp]
\caption{AdaSwarm \emph{vs.} other optimizers: The promising results of AdaSwarm using MSE (\emph{Differentiable Loss Function}) could address the sensitivity (\emph{to initialization}), and robustness (\emph{to multiple local minima}) in the classification dataset}
\begin{center}
\scalebox{0.95}{
 \begin{tabular}{|c|c|c|c|c|c|c|c|c|c|}
\hline
\textbf{Dataset} & \textbf{Metrics} & \multicolumn{8}{c|}{\textbf{Optimizer}} \\ \hline
 &  & {\color[HTML]{333333} SGD} & {\color[HTML]{333333} \begin{tabular}[c]{@{}c@{}}Emulation  of SGD \\ with PSO parameters\end{tabular}} & AdaGrad & AdaDelta & RMSProp & AMSGrad & Adam & AdaSwarm \\ \hline
 & Loss & 0.072 & 0.073 & 0.108 & 0.123 & 0.049 & 0.094 & 0.05 & \textbf{0.062} \\ \cline{2-10} 
\multirow{-2}{*}{Iris} & Accuracy & 92\% & 96.66\% & 88.22\% & 88.22\% & 98.22\% & 88.22\% & 98.22\% & \textbf{98.22\% \rpm 0.38\%} \\ \hline
 & Loss & 0.238 & 0.5 & 0.064 & 0.072 & 0.012 & 0.0167 & 0.0156 & \textbf{0.022} \\ \cline{2-10} 
\multirow{-2}{*}{Ionosphere} & Accuracy & 68.14\% & 95\% & 93.71\% & 92.57\% & 98.43\% & 98.14\% & 98.57\% & \textbf{98.48\% \rpm 0.28\%} \\ \hline
 & Loss & 0.198 & 0.124 & 0.127 & 0.1234 & 0.116 & 0.116 & 0.121 & \textbf{0.119} \\ \cline{2-10} 
\multirow{-2}{*}{Wisconsin Breast Cancer} & Accuracy & 80.79\% & 85.11\% & 84.20\% & 84.53\% & 85.34\% & 85.33\% & 84.82\% & \textbf{85.55\% \rpm 0.74\%} \\ \hline
 & Loss & 0.251 & 0.132 & 0.144 & 0.142 & 0.117 & 0.112 & 0.108 & \textbf{0.11} \\ \cline{2-10} 
\multirow{-2}{*}{Sonar} & Accuracy & 50\% & 81.00\% & 82.69\% & 81.97\% & 84.37\% & 86.05\% & 86.30\% & \textbf{86.28\% \rpm 0.34\%} \\ \hline
 & Loss & 0.219 & 0.2 & 0.162 & 0.203 & 0.119 & 0.137 & 0.127 & \textbf{0.13} \\ \cline{2-10} 
\multirow{-2}{*}{Wheat Seeds} & Accuracy & 66.67\% & 66.67\% & 78.70\% & 66.67\% & 85.23\% & 82.38\% & 81.75\% & \textbf{82\% \rpm 0.22\%} \\ \hline
 & Loss & 0.112 & 0.002 & 0.006 & 0.002 & 0 & 0 & 0 & \textbf{0} \\ \cline{2-10} 
\multirow{-2}{*}{Bank Note   Authentication} & Accuracy & 92.71\% & 100.00\% & 100.00\% & 100.00\% & 100.00\% & 100.00\% & 100.00\% & \textbf{100.00\% \rpm 0.0\%} \\ \hline
 & Loss & 0.461 & 0.187 & 0.509 & 0.4612 & 0.5312 & 0.48 & 0.18 & \textbf{0.19} \\ \cline{2-10} 
\multirow{-2}{*}{Heart Disease} & Accuracy & 53.87\% & 72.55\% & 47.81\% & 53.87\% & 46.12\% & 51.01\% & 74.07\% & \textbf{74.07\% \rpm 1.06\%} \\ \hline
 & Loss & 0.2365 & 0.195 & 0.174 & 0.172 & 0.176 & 0.169 & 0.17 & \textbf{0.169} \\ \cline{2-10} 
\multirow{-2}{*}{Haberman's Survival} & Accuracy & 73.70\% & 73.52\% & 75.49\% & 76.60\% & 76.79\% & 77.28\% & 76.96\% & \textbf{77.10\% \rpm 1.27\%} \\ \hline
\multirow{2}{*}{Wine} & Loss & 0.4 & 0.445 & 0.445 & 0.4 & 0.219 & 0.333 & 0.226 & \textbf{0.08} \\ \cline{2-10} 
& Accuracy & 59.92\% & 55.43\% & 55.43\% & 59.92\% & 66.67\% & 66.67\% & 76.78\% & \textbf{93.07\% \rpm 0.27\%} \\ \hline
 & Loss & 0.188 & 0.0718 & 0.094 & 0.096 & 0.0815 & 0.072 & 0.0918 & \textbf{0.07} \\ \cline{2-10} 
\multirow{-2}{*}{Car Evaluation} & Accuracy & 81.19\% & 91.46\% & 86.31\% & 86.15\% & 88.49\% & 90.10\% & 86.14\% & \textbf{92\% \rpm 0.19\%} \\ \hline
\multirow{2}{*}{Indian   Liver Patient} & Loss & 0.21 & 0.284 & 0.284 & 0.284 & 0.284 & 0.232 & 0.227 & \textbf{0.227} \\ \cline{2-10} 
 & Accuracy & 71.50\% & 71.50\% & 71.50\% & 70.72\% & 71.50\% & 72\% & 72.10\% & \textbf{72.63\% \rpm 0.55\%} \\ \hline
\multirow{2}{*}{Abalone} & Loss & 0.194 & 0.164 & 0.157 & 0.146 & 0.12 & 0.12 & 0.12 & \textbf{0.12} \\ \cline{2-10} 
 & Accuracy & 77\% & 77\% & 77\% & 79.24\% & 83.44\% & 83.00\% & 83.27\% & \textbf{83.54\% \rpm 0.26\%} \\ \hline
\multirow{2}{*}{Titanic} & Loss & 0.226 & 0.208 & 0.18 & 0.2 & 0.148 & 0.151 & 0.149 & \textbf{0.149} \\ \cline{2-10} 
 & Accuracy & 67.92\% & 70.59\% & 71.57\% & 70.26\% & 79.34\% & 80.39\% & 79.06\% & \textbf{80.53\% \rpm 0.41\%} \\ \hline
\multirow{2}{*}{Pima India Diabetes} & Loss & 0.307 & 0.2311 & 0.2411 & 0.285 & 0.1937 & 0.263 & 0.1908 & \textbf{0.193} \\ \cline{2-10} 
 & Accuracy & 66.73\% & 66.73\% & 64.51\% & 66.99\% & 71.67\% & 69.53\% & 70.76\% & \textbf{71.67\% \rpm 0.24\%} \\ \hline
\multirow{2}{*}{Agaricus Lepiota} & Loss & 0.24 & 0.16 & 0.08 & 0.03 & 0 & 0 & 0 & \textbf{0} \\ \cline{2-10} 
 & Accuracy & 91.63\% & 97.53\% & 97.46\% & 99.16\% & 100\% & 100\% & 100\% & \textbf{100\% \rpm 0.0\%} \\ \hline
\end{tabular}}
\label{table:adaswarmClassificationMSELoss}
\end{center}
\end{table*}

\begin{table*}[htbp]
\caption{AdaSwarm \emph{vs.} other optimizers: The promising results of AdaSwarm using MAE (\emph{Non-Differentiable Loss Function})}
\begin{center}
\scalebox{0.9}{
 \begin{tabular}{|c|c|c|c|c|c|c|c|c|c|}
\hline
\textbf{Dataset} & \textbf{Metrics} & \multicolumn{8}{c|}{\textbf{Optimizer}} \\ \hline
 &  & SGD & \begin{tabular}[c]{@{}c@{}}Emulation  of SGD \\ with PSO parameters\end{tabular} & AdaGrad & AdaDelta & RMSProp & AMSGrad & Adam & AdaSwarm \\ \hline
\multirow{2}{*}{Iris} & Loss & 0.22 & 0.177 & 0.319 & 0.195 & 0.138 & 0.226 & 0.224 & \textbf{0.211} \\ \cline{2-10} 
 & Accuracy & 77.77\% & 95.11\% & 66.67\% & 88.22\% & 88.22\% & 77.77\% & 77.77\% & \textbf{88.22\% \rpm 0.22\%} \\ \hline
\multirow{2}{*}{Ionosphere} & Loss & 0.468 & 0.34 & 0.295 & 0.32 & 0.216 & 0.218 & 0.217 & \textbf{0.217} \\ \cline{2-10} 
 & Accuracy & 64\% & 79.57\% & 74.42\% & 70.57\% & 79.4\%2 & 79.714\% & 79.428\% & \textbf{82.714\% \rpm 0.78\%} \\ \hline
\multirow{2}{*}{Wisconsin Breast   Cancer} & Loss & 0.385 & 0.411 & 0.349 & 0.348 & 0.349 & 0.349 & 0.349 & \textbf{0.334} \\ \cline{2-10} 
 & Accuracy & 65.10\% & 67.96\% & 65.10\% & 65.10\% & 65.10\% & 65.10\% & 65.83\% & \textbf{70.45\% \rpm 1.03\%} \\ \hline
\multirow{2}{*}{Sonar} & Loss & 0.498 & 0.245 & 0.304 & 0.275 & 0.21 & 0.199 & 0.1735 & \textbf{0.169} \\ \cline{2-10} 
 & Accuracy & 50\% & 83.65\% & 77.4\% & 81.49\% & 83.89\% & 87.74\% & 86.05\% & \textbf{89.53\% \rpm 0.33\%} \\ \hline
\multirow{2}{*}{Wheat Seeds} & Loss & 0.453 & 0.333 & 0.281 & 0.337 & 0.2722 & 0.333 & 0.26 & \textbf{0.245} \\ \cline{2-10} 
 & Accuracy & 64.92\% & 66.67\% & 75.87\% & 67.3\% & 74.6\% & 70.79\% & 74.032\% & \textbf{79.52\% \rpm 0.27\%} \\ \hline
\multirow{2}{*}{Bank Note   Authentication} & Loss & 0.273 & 0.021 & 0.0255 & 0.0049 & 0 & 0.001 & 0.002 & \textbf{0.001} \\ \cline{2-10} 
 & Accuracy & 79.07\% & 100\% & 100\% & 100\% & 100\% & 100\% & 100\% & \textbf{100\% \rpm 0.0\%} \\ \hline
\multirow{2}{*}{Heart Disease} & Loss & 0.45 & 0.39 & 0.461 & 0.461 & 0.449 & 0.461 & 0.389 & \textbf{0.378} \\ \cline{2-10} 
 & Accuracy & 55.21\% & 72.22\% & 53.87\% & 53.87\% & 54.88\% & 53.87\% & 62.96\% & \textbf{73.23\% \rpm 0.27\%} \\ \hline
\multirow{2}{*}{Haberman's Survival} & Loss & 0.264 & 0.264 & 0.264 & 0.264 & 0.264 & 0.264 & 0.264 & \textbf{0.264} \\ \cline{2-10} 
 & Accuracy & 73.52\% & 73.52\% & 73.52\% & 73.52\% & 73.52\% & 73.52\% & 73.52\% & \textbf{73.52\% \rpm 0.0\%} \\ \hline
\multirow{2}{*}{Wine} & Loss & 0.599 & 0.33 & 274 & 0.315 & 0.333 & 0.4 & 0.247 & \textbf{0.333} \\ \cline{2-10} 
 & Accuracy & 40.00\% & 66.67\% & 73.22\% & 69.10\% & 66.67\% & 59.92\% & 75.84\% & \textbf{66.67\% \rpm 1.23\%} \\ \hline
\multirow{2}{*}{Car Evaluation} & Loss & 0.261 & 0.221 & 0.15 & 0.15 & 0.118 & 0.15 & 0.15 & \textbf{0.138} \\ \cline{2-10} 
 & Accuracy & 85.01\% & 88.25\% & 85.01\% & 85.01\% & 88.10\% & 85.01\% & 85.01\% & \textbf{92.10\% \rpm 0.31\%} \\ \hline
\multirow{2}{*}{Indian Liver Patient} & Loss & 0.5 & 0.43 & 0.285 & 0.285 & 0.285 & 0.285 & 0.285 & \textbf{0.28} \\ \cline{2-10} 
 & Accuracy & 50\% & 57\% & 71.50\% & 71.50\% & 71.50\% & 71.50\% & 71.50\% & \textbf{72.10\% \rpm 0.18\%} \\ \hline
\multirow{2}{*}{Abalone} & Loss & 0.391 & 0.335 & 0.232 & 0.23 & 0.229 & 0.229 & 0.23 & \textbf{0.284} \\ \cline{2-10} 
 & Accuracy & 77\% & 77\% & 77\% & 77\% & 77\% & 77\% & 77\% & \textbf{82.91\% \rpm 0.38\%} \\ \hline
\multirow{2}{*}{Titanic} & Loss & 0.366 & 0.4 & 0.319 & 0.311 & 0.3 & 0.305 & 0.306 & \textbf{0.294} \\ \cline{2-10} 
 & Accuracy & 64.21\% & 68.20\% & 68.62\% & 69.46\% & 69.95\% & 69.74\% & 69.74\% & \textbf{82.00\% \rpm 0.12\%} \\ \hline
\multirow{2}{*}{Pima India Diabetes} & Loss & 0.348 & 0.348 & 0.349 & 0.348 & 0.348 & 0.348 & 0.348 & \textbf{0.348} \\ \cline{2-10} 
 & Accuracy & 65.10\% & 65.23\% & 65.10\% & 65.10\% & 65.10\% & 65.10\% & 65.10\% & \textbf{73.43\% \rpm 0.36\%} \\ \hline
\multirow{2}{*}{Agaricus Lepiota} & Loss & 0.16 & 0.066 & 0.058 & 0.06 & 0.052 & 0.052 & 0.052 & \textbf{0.07} \\ \cline{2-10} 
 & Accuracy & 85.96\% & 99.00\% & 94.53\% & 94.02\% & 94.72\% & 94.72\% & 94.72\% & \textbf{99.85\% \rpm 0.04\%} \\ \hline
\end{tabular}}
\label{table:adaswarmClassificationMAELoss}
\end{center}
\end{table*}


\section{AdaSwarm}
\label{adaswarm}
We propose a fast, gradient-free optimizer called AdaSwarm. The Adam \cite{adam} optimizer, is a very popular optimization algorithm used extensively in NNs, requires first-order derivatives and has little memory requirements. The Adam optimizer calculates adaptive learning rates for different parameters from the estimates of second and first moments of the gradients. AdaSwarm is based on the Adam optimizer but replaces the gradient with the computed, approximate gradients via the equivalence theorems, replacing
these gradients in Adam’s update rule. Experimental results show that AdaSwarm has a lower execution time and comparable (\emph{and most times superior}) performance to Adam.

\begin{algorithm}

\textbf{Require:} $\eta$: Learning Rate\;
\textbf{Require:} $\beta_1$, $\beta_2$ $\in$ [0, 1): Exponential decay rates for the moment estimates\;
\textbf{Require:} $f(\theta)$: Function with parameter $\theta$\;
\textbf{Require:} $\theta_0$: Initial parameter vector\;
$m_0$ $\leftarrow$ 0 (Initialize 1st moment vector)\;
$v_0$ $\leftarrow$ 0 (Initialize 2nd moment vector)\;
$t$ $\leftarrow$ 0 (Initialize timestep)\;
\While{$\theta_t$ not converged}{
    $t \leftarrow t + 1$\;
    $g_t$ $\leftarrow$ Approximates Gradients  (get gradients w.r.t. stochastic objective at timestep $t$)\;
    $m_t \leftarrow \beta_1 \cdot m_{t-1} (1 - \beta_1) \cdot g_t$ \;
    $v_t \leftarrow \beta_2 \cdot v_{t-1} + (1 - \beta_2) \cdot g^2_t$ \;
    $\theta_t \leftarrow \theta_{t-1} - \eta \cdot \frac{m_t}{\sqrt{v_t}+\epsilon}$}
 Return $\theta_t$
 \caption{AdaSwarm}
\end{algorithm}

Adam is a combination of SGD with momentum and RMSProp. It leverages the momentum by using the moving average of the gradient instead of the gradient itself like in SGD (SGD) with momentum and the squared gradients are used to scale the learning rate like in RMSProp. Besides these capabilities, when we add approximate gradients calculated using EMPSO (\emph{A fast converging Particle Swarm Optimizer}), this approach becomes a truly derivative-free optimizer.  AdaSwarm is thus a combination of RMSProp, SGD with Momentum and EMPSO with the aim of providing speed and acceleration to train NNs.

\subsection{Convergence of AdaSwarm}
As there is no meaningful size of a function for the optimization tasks (\emph{convex/non-convex}), in contrast to the GD algorithm for convex problems which finds the location of the minimum of a multi-variable function, we cannot clearly describe the computational complexity of AdaSwarm. However, there is an ample amount of literature that supports the convergence of Adam \cite{bock2018improvement}. Since AdaSwarm is based on the Adam optimizer, the algorithm is similar in structure and therefore we can infer whether it converges or not, based on the upper bounds for convergence rate of Adam.

The time complexity of AdaSwarm can be computed as follows: time taken to find gradient $\times$ steps required to reach the optimum. The time taken to find the gradient is the time complexity required to find the gradient. But since we approximate the gradient, for complex loss functions, the gradient calculation time taken by EMPSO is lower than mathematically computing it. Also, we note that the steps required to reach the optimum is lower for AdaSwarm. Thus, the AdaSwarm algorithm performs better/faster than Adam. The complexity analysis of EMPSO with more explanations is provided in Section 7 of the supp file.
 

\subsection{REMPSO: Rotation Accelerated EMPSO for high dimensional data}
\label{section:rempso}
PSO often fails in searching for the optima in the presence of a large number of variables mainly because of the computational cost required and the inability to escape a sub-plane of the whole search space. Hatanaka \cite{inbook} proposed a Rotated Particle Swarm Optimizer to improve the performance of PSO when dealing with high-dimensional optimization problems. The motivation for this work was to tackle the computational cost of EMPSO for solving multi-class classification problems. Using Rotated PSO, we are able to overcome this problem, increasing the convergence speed of \textbf{AdaSwarm} on CV data sets (\emph{see Section 5 of the supp file for REMPSO details and results}). REMPSO is a redefined version of EMPSO by leveraging the Rotation acceleration property to solve high-dimensional optimization problems in classification. 
    

\section{Experiments}
\label{section:experiments}

\subsection{Experimental Setup}

For testing AdaSwarm in NNs \footnote{\url{https://github.com/rohanmohapatra/torchswarm} --- EMPSO pytorch}  \footnote{\url{https://github.com/anuwu/PSO-Grad} --- EMPSO numpy}, it was compared with respect to several state-of-the-art optimizers and applied on several benchmark data sets. 
The corresponding results are presented in \textbf{Table~\ref{table:adaswarmClassificationMSELoss} and in Table~\ref{table:adaswarmClassificationMAELoss}}.
In these experiments, for each of the data sets previously mentioned, we have kept the parameters of EMPSO constant (\textbf{$c_1 = 0.8$, $c_2 = 0.9$, $\beta = 0.9, \eta=0.1$}). The choice of $c_1,c_2$ was guided by detailed stability analysis (See the supp file, Section 3).  
For consistency, a NN with one hidden layer with 10 neurons was used. The results shown are therefore unbiased and kept at default values to show the speed of convergence of AdaSwarm. 

\subsection{True Gradient and Approximate Gradient Comparison}
The gradients were calculated for the functions mentioned in Tables \ref{table:gdVsEmulatedGD}-\ref{table:sgdVsEmulatedSGDEMPSONonDiff}. We plotted the approximate gradient values computed using the PSO parameters defined in Section~\ref{section:equivalenceSGDEMPSO} and the true gradients. The results are shown in Fig.~\ref{fig:trueVsApproxGradient} and Fig.~\ref{fig:trueVsApproxGradientNonDiff}. This graph shows that the use of this approach for computing gradients (\emph{for differentiable and non-differentiable functions}) is indeed a viable choice and, consequently it can be embedded in the backpropagation algorithm. 




\subsection{AdaSwarm vs. Optimizers for Classification Datasets}

Using the approximated derivative, we replace the backward pass in the loss to this derivative. With such a modification, every time the algorithm backpropagates, it uses the custom approximate gradients throughout the layer to update the weights. A new loss was defined, in the forward pass. The binary cross-entropy loss was computed and in the backward pass, we replaced the gradient with our approximated gradient. The data set was divided into batches. Then, for an epoch, the batch loss and the accuracy were calculated using the optimizer. The losses were running averages in the batch. The same running averages were applied to the accuracies. The epoch loss, average accuracy with standard deviation were reported. For comparison purposes, we adopted SGD, SGD emulation with PSO parameters, AdaDelta \cite{adadelta}, AMSGrad \cite{amsgrad}, RMSProp, Adam, and AdaSwarm on data sets from the UCI repository \cite{Dua:2019} (\emph{Section 4 of the supp file}). 
AdaSwarm's iterations to converge were compared to those of Adam by keeping the accuracy as a threshold (\emph{see Tables 6, 7 \& 8 in the supp file}). 
AdaSwarm consistently outperformed Adam across all data sets (\emph{sometimes AdaSwarm is twice as fast as Adam}). 
\subsection{Convolutional NN (CNN) : Architecture used for comparisons between Adam and AdaSwarm}
We define a simple CNN architecture for conducting tests on benchmark  data sets from computer vision. 
This model contains 2 convolution layers having a $3 \times 3$ kernel, a max-pooling layer and a flattened layer connected to a 128 neuron hidden layer, with an output layer of 10 neurons for 10 classes. 
The total weights for the MNIST model (\emph{say}) sums up to 1,199,882 having a $28 \times 28$ image size. 
As the image size varies for CIFAR-10, the weights will vary, but the architecture remains the same (\emph{See Table 10 of the supp file for results}). 
To further validate AdaSwarm, it was extended to deeper NNs (\emph{considered as benchmarks in image classification}) like ResNet-18 \cite{he2015deep}. The results are presented in Table 11 of the supp file. 

\noindent \textbf{Hardware Considerations:}
AdaSwarm requires lower number of epochs, cheaper architecture (\emph{by comparing the hardware requirements of ResNet18 and ResNet 50}) and lower CPU/memory utilization. 
Even for the best hardware processing units \cite{gpu}, 
running an Epoch in ResNet50 requires almost twice the time required by ResNet18 (Section 3 of the supp file). 


\section{Discussion}
\label{section:discussions}
NN computation can be viewed as a sequence of functional compositions \cite{shenslides} \cite{Shen_2019}. This is one way to understand how nodes in hidden-layers progressively represent \emph{more complex} features. However, sufficiently deep networks carry significant computation overhead in the training process. Also, lack of theoretical guarantees makes us wonder if
there is some theoretical validation on how deep networks can be approximated by shallower ones. AdaSwarm constitutes an empirical validation of this hypothesis, having solved the benchmark problems under resource constraints, in a reasonably parsimonious fashion \cite{yedida2019lipschitzlr}\cite{sridhar2020parsimonious}. 

The proposed approach considers, at each iteration, that $c_1 r_1, c_2 r_2$, are computed by taking the average of $n$ particles and $v^{(t-1)}$ is taken of the particle that influences the gbest. $c_1$ and $c_2$ are parameters typically used in PSO. 
We usually adopt the $n$ particles of PSO to obtain the values and substitute them. 
That is why $c_1r_1 + c_2r_2$ is averaged over $n$ particles whereas $r_1$ and $r_2$ change per particle, drawn from $U(0,1)$.
The global best is  governed by the velocity of one particle whose velocity is taken to 
be $v^{(t-1)}$. No additional tuning of parameters was required on the data sets\footnote{Implementation: \url{https://github.com/rohanmohapatra/pytorch-cifar}}. 

The simple yet elegant AdaSwarm model helped us to solve  benchmark problems (\emph{different papers provide different benchmarks}) quite efficiently. Since AdaSwarm is an optimizer, state-of-the-art 
approaches used to validate it must be optimizers rather than models. We illustrate the scalability of our proposed approach 
for computer vision data sets, on ResNet 18 architecture in which our approach was able to outperform other SOTA optimizers \cite{hosseini2020adas} tested on ResNet 18. However, we went ahead and checked the performance of these (Adam, Diffgrad etc) on ResNet 50 from\cite{diffgrad} and our AdaSwarm on a stingier architecture ResNet 18 outperformed.
It is worth noting that MAE is an unusual choice for a 
GD training method until recently \cite{saha2020lalr} and is approximated well.


Momentum in PSO (i.e., EMPSO) is equivalent to a history of all velocities in an exponentially weighed fashion, making the search curve smoother since it contains all the momentum history. However, memory management does not have an adverse impact as we are not storing any velocities. 
Instead, it is a cumulative sum in the sense that when a new velocity is computed, it's accumulated over time, helping us reach the solution around the optimum. Inertia ceases to exist near the optimum. We have also used the bias correction in AdaSwarm (\emph{default implementation}). 
The authors in \cite{Ye2013} \cite{Qian2018} claim that PSO converges near the optimum and by using initial and middle phases of the particles, we can factor in the velocity of the particles in the equivalence to go ahead with the approximation.

A subtlety to be noted in our gradient approximation formula is its applicability in the event of the underlying theoretical assumptions not being met. Say that a minority of the particles converged at $g_{best}$. Our gradient formulas are still applicable because of its first order form $\kappa(w-g_{best})$ --- \emph{for an appropriate constant $\kappa$}. What ultimately matters is that the best possible $g_{best}$ be found, and not that the entire swarm converge to it. The purpose of our derivations is to \emph{legitimize} the usage of a first-order approximation to the gradient, from PSO parameters, without making an assumption on the differentiability of the objective function. Essentially, it is a connection between swarm intelligence and gradient-based optimization (\emph{two separate fields on their own right}). Our theoretical assumptions are not only conducive to the derivation, but also offer substantial simplification in terms of bare understanding. The \emph{usage} of the formula does not distinguish between complete and partial convergence of the swarm so long as it converges at the correct $g_{best}$.

Moreover, we have adequately justified in aforementioned references that the discovery of a sub-optimal $g_{best}$ does not make the NN weights (ultimately optimizing), via backpropagation, sub-optimal. We \emph{do not} claim, however, that EMPSO approximated gradients be used in other contexts, where it is crucial that the swarm find the global optima of the objective function with guarantee and reliability. It is anticipated that approximated and true gradients will only deviate largely for query points $w$ far from $g_{best}$. Thus a blind replacement of true gradients with EMPSO approximation gradients, in an arbitrary optimization context, for $w$ far from $g_{best}$, would be unwise. The sole purpose of deriving formulas approximating gradients, and its subsequent use in AdaSwarm, is a means of overcoming the non-differentiability of loss functions in the \emph{specific} context of NNs, which we have successfully achieved and demonstrated by experiments.

Tables \ref{table:gdVsEmulatedGD}, \ref{table:sgdVsEmulatedSGDEMPSO}, \ref{table:sgdVsEmulatedSGDEMPSONonDiff} run PSO and EMPSO on sample 1-D functions and the true global minima is reliably obtained by the swarm. In some instances, despite the swarm partially converging, the correct local/global minima is obtained because it is only required that one particle find the $g_{best}$. Moreover table 4 in the supp file shows negligible deviation between true and approximated gradients for $w$ far from the $g_{best}$. 


\section{Conclusions and Future Work}
A theoretical relation directly connecting the gradient to the parameters of PSO had never been presented. 
The derivative of any function with countably finite singularities may now be precisely expressed in terms of the PSO and EMPSO parameters. Furthermore, the approximation of derivatives is seamlessly extended to approximate error gradients in the backpropagation algorithm used in NNs. our method surpasses standard PSO and SOTA in terms of speed of convergence and accuracy. A new optimizer, AdaSwarm, based on the widely used Adam optimizer, is introduced and tested against Adam in training NNs. 
The proposed 
incentivizes the use more natural (\emph{like MAE}), non-differentiable loss functions in NNs.

Deep MLPs \cite{domingos2020model} learn features in the gradient space via path kernels. 
Therefore, if the GD does not do well, the model will have a poor performance. 
This leads us to ponder whether AdaSwarm learns better representations. 
What is the equivalent of a path kernel here? 
Does AdaSwarm discover better paths and also better approximations of the gradients (\emph{for non-differentiable, rugged landscapes})? 
Can we frame kernels as the integral (\emph{over the path of gradient trajectory}) of the dot product between gradients (\emph{of the $x$ with $x_i$}) and argue that such kernels learn better representations of the gradients? 
Thus, seeing optimization via AdaSwarm just as a means to fit the model can be extended 
to making it integral to the learning process itself. 

As an advancement to the PSO gradient equivalence theorem, it would be worthwhile to develop an analogous proof for Hessian approximations using PSO for differentiable and non-differentiable functions. These are questions worth pursuing in future research.


\section*{Acknowledgements}
Snehanshu Saha would like to thank the SERB-DST, Govt.of India [SERB-EMR/ 2016/005687] and RIG from BITS Pilani K K Birla Goa Campus for supporting this research and C.A.C. Coello acknowledges support from
CONACyT project no. 1920 and from a 2018 SEP-Cinvestav grant (application no. 4). 
He is also partially supported by the Basque Government through the BERC 
2018-2021 program by the Spanish Ministry of Science.

\ifCLASSOPTIONcaptionsoff
  \newpage
\fi

\bibliographystyle{IEEEtran}
\bibliography{main}

\end{document}